\pdfoutput=1
\documentclass{article}

\usepackage{arxiv}

\usepackage{inputenc}
\usepackage{fontenc}
\usepackage{hyperref}
\usepackage{url}
\usepackage{booktabs}
\usepackage{amsfonts}
\usepackage{nicefrac}
\usepackage{microtype}
\usepackage{multirow}
\usepackage{booktabs}
\usepackage{makecell}
\usepackage{lipsum}
\usepackage{graphicx}
\usepackage{amsmath}

\title{SatVision-TOA: A Geospatial Foundation Model for Coarse-Resolution All-Sky Remote Sensing Imagery}

\author{
 Caleb S. Spradlin,$^{1,2}$ Jordan A. Caraballo-Vega,$^{1}$ Jian Li,$^{1,2}$ Mark L. Carroll,$^{1}$
 \\
 \textbf{Jie Gong,$^{3}$  Paul M. Montesano,$^{1,4}$}
}

\date{}

\begin{document}
\maketitle
\vspace{0.2in}
\begin{center}
    \normalsize{$^{1}$Data Science Group, NASA Goddard Spaceflight Center}
    \\
    \normalsize{$^{2}$InuTeq, LLC.}
    \\
    \normalsize{$^{3}$Climate and Radiation Lab, NASA Goddard Spaceflight Center}
    \\
    \normalsize{$^{4}$ADNET Systems, Inc.}

\end{center}
\vspace{0.4in}
\begin{abstract}
Foundation models have the potential to transform the landscape of remote sensing (RS) data analysis by enabling large computer vision models to be pre-trained on vast amounts of remote sensing data.  These models can then be fine-tuned with small amounts of labeled training and applied to a variety of mapping and monitoring applications. Most existing foundation models are designed for high spatial resolution, cloud-free satellite imagery or photos, limiting their applicability in scenarios that require frequent temporal monitoring or broad spectral profiles. As a result, foundation models trained solely on cloud-free images have limited utility for applications that involve atmospheric variables (e.g., cloud or aerosol) or require atmospheric corrections. We introduce SatVision-TOA, a novel foundation model pre-trained on 14-band MODIS L1B Top-Of-Atmosphere (TOA) radiance imageries, addressing the need for models pre-trained to handle moderate- and coarse-resolution all-sky remote sensing data. The SatVision-TOA model is pre-trained using a Masked-Image-Modeling (MIM) framework and the SwinV2 architecture, and learns detailed contextual representations through self-supervised learning without the need for labels. It is a 3 billion parameter model that is trained on 100 million images. To our knowledge this is the largest foundation model trained solely on satellite remote sensing imagery. Initial results indicate that SatVision-TOA achieves superior performance over baseline methods when applied to downstream tasks such as 3D cloud retrieval. Notably, the model achieves a mean intersection over union (mIOU) of 0.4638, a substantial improvement over the baseline mIOU of 0.218. Additionally, the rate of false negative results in the fine-tuning task were reduced by over 50\% compared to the baseline. This improvement in mIOU and sensitivity demonstrates the model's enhanced capability to delineate segmented regions effectively for this specific task. Our work advances pre-trained vision modeling for multispectral remote sensing by learning from a variety of atmospheric and aerosol conditions to improve cloud and land surface monitoring.
\end{abstract}

\textit{The copyright holder for this preprint is the author/funder, who has granted arXiv a license to display the preprint in perpetuity. This article is a US Government work. It is not subject to copyright under 17 USC 105 and is also made available for use under a CC0 license.}

\section{Introduction}
\subsection{The potential of foundation models in remote sensing}
Foundation models have emerged as powerful tools in remote sensing data analysis, providing the ability to pre-train large computer vision models on vast datasets for use in diverse downstream tasks. However, most of these models have focused on high spatial resolution satellite imagery, optimizing for tasks like fine-grained object detection, segmentation, (buildings, vehicles, airplanes, etc.), field-scale land cover classification, and change detection \cite{cong2022satmae, jakubik2310foundation, reed_scale-mae_2023}. High spatial resolution data offers detailed spatial information but is often constrained by limited temporal frequency, making it less suitable for applications that may require repeat observations (intra-seasonal agriculture assessment, fine-scale phenology, etc.) as well as limited spectral bands (often only Red, Green, Blue and NIR). Compared to high-resolution imagery, moderate- and coarse-resolution data, such as Top-Of-Atmosphere (TOA) imagery from instruments such as the Advanced Baseline Imager (ABI) on the Geostationary Operational Environmental Satellites (GOES-R), the Visible Infrared Imaging Radiometer Suite (VIIRS), and the MODerate-resolution Imaging Spectroradiometer (MODIS), provides broader coverage and higher revisit rates. This allows for more frequent monitoring of atmospheric and land surface phenomena.

\subsection{Leveraging frequent multispectral spaceborne observations} 
Currently, a suite of spaceborne multispectral sensors offer frequent (daily) coverage along a broad spectral range at a moderate resolution. MODIS is a passive spectroradiometer on board the Terra and Aqua satellites. Flying at different equator crossing times, Terra-MODIS (morning orbit) and Aqua-MODIS (afternoon orbit) together revisit any place on Earth every day, acquiring data in 36 spectral bands ranging from visible to thermal-infrared (TIR) wavelengths with 250 m spatial resolution for Band 1 and 2 at nadir to 1 km spatial resolution for thermal bands. These observations help in monitoring environmental changes, atmospheric phenomena and other diverse applications. The MODIS successors include the Visible Infrared Radiometer Suite (VIIRS) onboard NOAA’s polar-orbiting JPSS satellite series, and SEVIRI onboard European’s polar-orbiting Meteosat series.  In addition, there are spectrometers on geostationary platforms such as the Advanced Baseline Imager (ABI) onboard NOAA’s GOES geostationary satellite series, and the Atmospheric Humidity Imager (AHI) onboard Japan’s Himawari geostationary platforms. Because of the data rate consideration (images every 10 – 20 minutes), ABI and AHI have only 16 channels, 14 of which can find similar wavelengths with 14 MODIS bands as described in Table \ref{tab:modis_bands}, but they can provide a much higher temporal sampling rate which enables weather monitoring. ABI data have been extensively cross compared with MODIS for similar bands to demonstrate consistency \cite{liang_preliminary_2016, chang_assessment_2019}. 

\subsection{Vision transformers can improve with more frequent observations} 
Machine learning has been used in many forms to analyze satellite remote sensing data for decades. Until recently, convolutional neural networks (CNN) \cite{he_deep_2015, krizhevsky2012imagenet, simonyan2014very} and fully convolutional networks (FCN) \cite{sermanet2013overfeat, long2015fully} have been the de-facto model of choice for many computer vision tasks over the past decade \cite{caraballo2023optimizing, zhu2017deep}. Recently, following the remarkable success achieved using Transformer models in natural language processing (NLP) \cite{vaswani_attention_2023}, the development of Vision Transformers (ViTs) was adapted to computer vision tasks such as image classification, semantic segmentation and object detection. Dosovitskiy et al. \cite{dosovitskiy_image_2021} applied a standard Transformer directly to images by splitting the image into a sequence of 16$\times$16 subsets referred to as patches, not focusing on pixels, then input to the Transformer the sequence of embeddings for those patches. The image patches were treated as tokens in NLP applications. These models led to state-of-the-art results on the ImageNet dataset. The ViT model architecture is highly scalable and benefits from large amounts of data, effectively utilizing extensive datasets while maintaining a low propensity for over-fitting. This scalability allows ViTs to improve performance as more data becomes available. Vision transformers can take advantage of frequent multispectral observations. When trained on a large volume of data (millions of images), ViT shows superior accuracy, compared with state-of-the-art CNNs \cite{dosovitskiy_image_2021}. The main driving force behind the ViT is the multi-head self-attention mechanism. It helps ViT to derive long-range contextual dependencies between pixels in images \cite{park2022vision}.  

\subsection{Remote sensing foundation ViT models capture spatial patterns} 
Foundation models (FMs) are typically large-scale pre-trained models which can be used as a starting point to fine-tune the model for a downstream task. These foundation models are trained on large amounts of remote sensing data, such as satellite imagery, using techniques such as self-supervised learning (SSL). Through the capture of important spatial and spectral information in data, FMs can be fine-tuned for specific tasks, including land cover classification and cloud detection. The core idea is that with extensive pre-training, the foundation model learns the underlying patterns and characteristics of the input data, even in the absence of explicit labels or a defined mapping task. This approach enables downstream users to fine-tune the pre-trained model with a limited number of labeled examples, hence reducing the computational cost required to address specific scientific questions. Users can apply their own labels to fine-tune the foundation model for their specific tasks, and, in theory, achieve greater computational efficiency since the model has already learned the intrinsic relationships in the input data; and only needs to adapt to the task-specific label assignment.

Remote sensing FMs have been fueled by the ability to leverage large amounts of unlabeled data through SSL techniques. These techniques impart learning through robust grouped representations based on similarities within the data without the need for labels. Two main approaches are used: [1] \textit{contrastive learning schemes} learn by comparing correlated views of the same image, while [2] \textit{predictive coding}, analogous to the masked language model used for training NLP models, trains models to predict missing parts of the input image from the observed parts. Examples of these models include SatMAE \cite{cong2022satmae}, which pre-trains transformers on temporal and multi-spectral high-resolution satellite imagery, and Privthi-HLS \cite{jakubik2310foundation}, pre-trained on 30 m Harmonized Landsat Sentinel scenes \cite{masek2021hls}. Scale-MAE \cite{reed_scale-mae_2023} pushes the boundaries further with scale-aware masked auto-encoding for multi-scale geospatial representation learning. These models have excelled in tasks such as image classification, change detection, and semantic segmentation, particularly using high spatial resolution satellite imagery because of their ability to represent spatial patterns. While these models perform well in high spatial resolution settings, there remains a critical need for FMs that are tailored to moderate- to coarse-resolution data with broad spectral ranges and high temporal frequency. Many vital remote sensing tasks-such as cloud reconstruction, environmental monitoring, and disaster management- depend on data with higher temporal resolution and broad spectral coverage.

\begin{table}
    \centering
    \begin{tabular}{ ccccc } 
        \hline
        Index & Band & Primary Use & Resolution & Central Wavelength \\
        \Xhline{3\arrayrulewidth}
        0 & 1 & Land/Cloud/Aerosols Boundaries & 250m & 0.659 $\mu$m \\ 
        1 & 2 & Land/Cloud/Aerosols Boundaries & 250m & 0.865 $\mu$m\\ 
        2 & 3 & Land/Cloud/Aerosols Properties & 500m & 0.47 $\mu$m\\ 
        3 & 6 & Land/Cloud/Aerosols Properties & 500m & 1.64 $\mu$m\\ 
        4 & 7 & Land/Cloud/Aerosols Properties & 500m & 2.13 $\mu$m\\ 
        5 & 21 & Surface/Cloud Temperature & 1km & 3.96 $\mu$m\\ 
        6 & 26 & Cirrus Clouds, Water Vapor & 1km & 1.375 $\mu$m\\ 
        7 & 27 & Cirrus Clouds, Water Vapor & 1km & 6.72 $\mu$m\\ 
        8 & 28 & Cirrus Clouds, Water Vapor & 1km & 7.33 $\mu$m\\ 
        9 & 29 & Cloud Properties & 1km & 8.55 $\mu$m\\ 
        10 & 30 & Ozone & 1km & 9.73 $\mu$m\\ 
        11 & 31 & Surface/Cloud Temperature & 1km & 11.03 $\mu$m\\ 
        12 & 32 & Surface/Cloud Temperature & 1km & 12.20 $\mu$m\\ 
        13 & 33 & Cloud Top Altitude & 1km & 13.34 $\mu$m\\ 
        \hline
    \end{tabular}
    \vspace{0.5cm}
    \caption{Description of the 14 MODIS spectral bands used in pre-training the SatVision-TOA model. This table highlights the spectral properties of MODIS channels, spanning visible, near-infrared, and thermal infrared wavelengths. These channels were chosen because they closely align with 14 of the 16 GOES ABI channels, enabling comprehensive pre-training for remote sensing tasks requiring diverse spatial and spectral knowledge. The MODIS instrument, aboard Terra and Aqua satellites, provides global coverage with varying spatial resolutions to support environmental and atmospheric monitoring applications.}
    \label{tab:modis_bands}
\end{table}

\section{Objectives} 
In this paper we introduce SatVision-TOA, a foundation model pre-trained on 14-band MODIS TOA imagery \cite{modis_science_team_mod021km_2017}.  SatVision-TOA leverages the spectral diversity of moderate-resolution TOA imagery, using all-sky data that captures natural atmospheric variability, including cloud cover, without relying on cloud-cleared images. This approach expands the scope of foundation models in remote sensing to tasks requiring rich spatial and spectral knowledge. By pre-training on MODIS TOA data, which includes 14 spectral channels spanning visible, near-infrared, and thermal-infrared wavelengths as presented in Table \ref{tab:modis_bands}, the model develops robust contextual representations needed for atmospheric science and environmental monitoring. The work on the pre-trained SatVision-TOA model has been in the spirit of open-science to enhance the use of machine learning within the expanding remote sensing community.

\section{Methodology}
\subsection{Developing a remote sensing pre-training dataset with MODIS TOA} We have developed an extensive pre-training dataset comprised of TERRA-MODIS Top-Of-Atmosphere (TOA) image chips to facilitate the pre-training of SatVision-TOA model. Each chip is a 128$\times$128 pixel image subset of a global daily composite made from MODIS TOA all-sky swaths. Composites were generated for approximately 270 days per year from 10 different years of the MODIS record, March 2000 to present.  This provides a data record that spans all seasons across multiple decades, which allows us to create a robust sample spatially and temporally to use in model training.  Scaling experiments were conducted to assess both model size (600 million to 3 billion parameters) and number of image chips in the training set (2 million to 100 million), these results are shown in the Appendix. For the final model we used the SwinV2 Giant architecture with 3 billion parameters and 100 million pre-training chips generated from global 1 km composites of MODIS Level 1B (L1B) TOA data.

\subsubsection{Preparing daily TOA composites from swath data } 
MODIS data comes in three spatial resolutions 250 m, 500 m and 1 km where bands 1 and 2 are natively 250 m, bands 3 – 7 are natively 500 m and bands 8 – 36 are natively 1 km.  For this work all bands need to be at the same spatial resolution, so the finer resolution bands 1 – 7 have been aggregated to 1 km resolution by the Elliptical Weighted Averaging (EWA) method \cite{gumley2003creating, greene1986creating}.  With MODIS instrument geometry the image distortion and footprint size increase at off-nadir views, hence wherever there were overlapping observations we selected the observation with the lowest view zenith angle to maximize the fidelity of the observations.  Additionally, we only retained observations with solar zenith angles lower than 72° to avoid pixels with low solar illumination conditions which can lead to erroneous model predictions.

The MODIS TOA data originates from MODIS MOD02 Level 1B swaths, these contain calibrated and geolocated radiances for 36 bands. The first step in preparing the dataset involved compositing the 5-minute MODIS swaths into daily global composites at 1 km spatial resolution. This process aggregates the continuous swath data into a consistent daily grid. Next, the digital numbers (DNs) from the MODIS swaths were converted into TOA reflectance for visible and NIR bands.

\begin{equation}
    \text{Reflectance} = (DN - \text{Reflectance Offsets}) \times \text{Reflectance Scales} \times 100
\end{equation}

The calibration of brightness temperature ($BT$) values for TIR bands is a more complex process. It requires methods outlined in the MODIS Level 1B product user guide \cite{modis_software_l1b} and is implemented through the SatPy Python package \cite{hoese2019satpy}. This calibration step converts Digital Number values into physical meaning and units. Specifically, for TIR bands, the process involves using physical constants and the effective central wave-number ($WN$) to accurately derive brightness temperature.

\begin{equation}
    \text{Radiance} = (DN - \text{Radiance Offsets}) \times \text{Radiance Scales}
\end{equation}
\begin{equation}
    BT = \frac{c_2}{\text{WN} \times \ln\left(\frac{c_1}{\text{Radiance} \times \text{WN}^5} + 1\right)}
\end{equation}
\begin{equation}
    BT = \frac{(BT - tci)}{tcs}
\end{equation}

$c_1$ and $c_2$ are derived constants based on the Planck constant $h$, the speed of light $c$, and the Boltzmann constant $k$. $tcs$ is the temperature correction slope, and $tci$ is the temperature correction intercept.

For both the reflectance and the brightness temperature, the values were scaled to a data range of 0-1. This is essential to ensure that the values fall within a range suitable for machine learning (ML) models, improving convergence and stability during training. For reflectance, values were scaled by a factor of 0.01, effectively transforming the data range from 0-100 to 0-1. For brightness temperature, values were min-max scaled to fit within a 0 to 1 range. This was done by identifying the global minimum and maximum brightness temperature in the dataset and applying the standard min-max scaling formula.

\begin{equation}
    \text{Scaled Value} = \frac{\text{Original Value} - \text{Min}}{\text{Max} - \text{Min}}
\end{equation}

This pre-processing step ensured that both the reflectance and temperature features were normalized into a similar range, which is crucial for stable performance of deep learning models \cite{ioffe2015batch}.

\subsubsection{Devising a sampling strategy for characterizing land- and cloud-cover diversity}
To ensure a geographically diverse training data set, the image chip selection was stratified across each quadrant of global composites generated from the MODIS L1B TOA swaths (NW, NE, SW, SE). We sampled from a broad range of global cover types, including ocean extents, the full suite of terrestrial land cover, as well as cloud and not-cloud samples. Because only daytime TERRA-MODIS images are used for training this model, it is expected to have some limitations in downstream tasks that involve diurnal variations (e.g., cloud life cycle). Nevertheless, the downstream task example given in this paper demonstrates that sampling a fixed local time does not introduce significant caveats.  

A sampling strategy that provides thorough characterization across the diverse set of land- and cloud-cover types is key to the development of a robust ViT.  As such, even a robust sample of chips based on a purely geographic random sample will be insufficient due to variations in cover type probabilities \cite{ramezan2021effects, shetty2021assessing, aleissaee2023transformers}. This approach would under-sample rare cover types and over-sample those that are common. To avoid these biases in cover type sampling, we implemented a clustering method to sample based on land- and cloud-cover structure represented in the image chips. This method ensures a sampling of diverse image chips for pre-training.  

\subsection{Configuring a model architecture and pre-training} 
We utilized the Masked-Image-Modeling (MIM) framework \cite{xie_simMIM_2022} for pre-training SatVision-TOA. MIM, inspired by Masked Language Modeling \cite{aharoni2019massively} from Natural Language Processing, involves randomly masking patches of the input satellite imagery chip and training the model to predict missing regions. This approach allows the model to learn detailed contextual representations through self-supervision, which is particularly beneficial when labeled data is scarce.

The MIM framework was chosen for its ability to capture both local and global context \cite{khan2024survey} of the imagery and its features by reconstructing masked portions of the image chips. This process enhances SatVision-TOA’s understanding of visual patterns and structures serving as a robust foundation for remote sensing downstream tasks.

We implemented this framework using the Swin-Version 2 \cite{liu_swin_2022} model architecture, leveraging its hierarchical approach to portion the image chips into smaller overlapping windows and its efficient attention mechanisms. During the pre-training phase, 8$\times$8 patches were randomly masked from the image chips and the model was trained to predict the original pixel values of the masked regions. This pre-training, conducted on a 100 million set of image chips, enabled the model to learn the complex relationships between the spectral bands in the image chips. 

\subsection{Assessing model performance}
To validate the model we focused on two key applications: image reconstruction performance, which evaluates the masked-image-modeling task on unseen data, and 3D cloud retrieval, a downstream task targeting the prediction of vertical cloud structure.

For image reconstruction, the primary performance validation metric used in our experiment is the Structural Similarity Index Measure (SSIM) \cite{wang2004image}. Unlike traditional metrics such as Mean Squared Error (MSE), which focuses on pixel-wise absolute error, SSIM evaluates image quality by quantifying perceived changes in structural information. This makes SSIM more suitable for measuring performance in tasks where the goal is not only to reconstruct pixel intensity but also to preserve edges and textures.

For 3D cloud retrieval, we validate the model's performance using both the mIOU metric and pixel-wise accuracy. Accuracy assesses how closely the model's predictions align with the ground truth on a pixel-by-pixel basis, while mIOU evaluates the model's ability to delineate segmented regions and accurately capture cloud structures. Relying solely on pixel-wise accuracy can be problematic for imbalanced datasets, such is the case with the 3D cloud retrieval task, where non-cloud pixels outnumber cloud pixels. To address this limitation, mIOU provides a more robust measure by calculating the overlap between predicted and ground-truth vertical cloud masks, averaged across the dataset.

A summary of the findings from the experiments associated with these two applications that use the best SatVision-TOA configuration is presented in the Results section. Additionally, we examined model scalability to explore the effects of increasing model complexity and pre-training dataset size. The detailed results of model scalability are provided in Appendix Tables \ref{tab:ssim_score_configs}, \ref{tab:3dcloud_retrieval_table_config}, \ref{tab:l1_loss}.

\section{Results}
\subsection{Image Reconstruction}

\begin{figure}[t]
    \centering
    \includegraphics[scale=0.4]{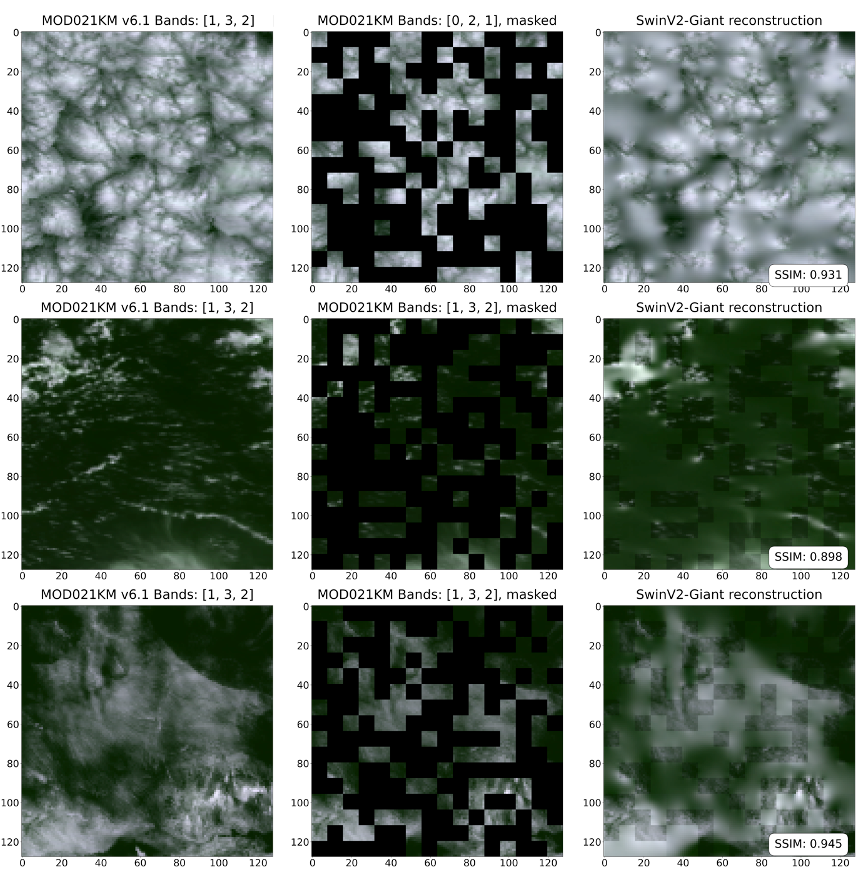}
    \caption{Examples of image reconstruction by SatVision-TOA. \textbf{Left:} MOD021KM v6.1 cropped image chip using MODIS bands [1, 3, 2]. \textbf{Middle:} The same images with randomly applied 8$\times$8 mask patches, masking 60\% of the original image. \textbf{Right: } The reconstructed images produced by the model, along with their respective Structural Similarity Index Measure (SSIM) scores. These examples illustrate the model's ability to preserve structural detail and reconstruct heterogeneous features, such as cloud textures and land-cover transitions, with high fidelity.}
    \label{fig:viz}
\end{figure}

To validate the reconstruction performance we measure how well SatVision-TOA could reconstruct masked inputs, a task consistent with the pre-training objective of masked-image modeling. Specifically, we evaluate how accurately the pre-trained model predicts masked regions of the input data based on the learned representations. The goal is to evaluate how well the model can reconstruct complicated and heterogeneous features such as different cloud- and land-cover types.

We used a carefully selected test dataset that represents a wide range of cloud- and land-cover types. This dataset consisted of 128 MODIS-TOA image chips which were withheld from the pre-training dataset, each with a randomly assigned mask. SatVision-TOA-Giant pre-trained with 100 M image chips achieves a SSIM score of 0.9289. Additional results comparing different model configurations, and their reconstruction performance is found at Appendix Table \ref{fig:ssim_channel}, \ref{tab:ssim_score_configs}.

\subsection{3D Cloud Retrieval Downstream Task}

\begin{table}[t]
    \centering
    \renewcommand{\arraystretch}{1.1}
    \resizebox{0.49\textwidth}{!}{
    \begin{tabular}{cccc}
    \Xhline{3\arrayrulewidth}
        Backbone & Pre-Train Dataset & mIOU & Accuracy \\
        \Xhline{3\arrayrulewidth}
        FCN Baseline & N/A & 0.2185 & 0.9320\\\hline
        SVTOA-Giant & 100M & 0.4638 & 0.9574\\\hline
        \Xhline{3\arrayrulewidth}
    \end{tabular}
    }
    \vspace{0.1cm}
    \caption{\label{tab:3dcloud_retrieval_table} Results of 3D cloud retrieval using various backbone models. The table compares the mean Intersection over Union (mIOU) and accuracy achieved by the FCN baseline and the SVTOA-Giant model pre-trained on a dataset of 100 million images. The SVTOA-Giant demonstrates a significant improvement in performance, achieving an mIOU of 0.4638 and accuracy of 0.9574 compared to the FCN baseline's mIOU of 0.2185 and accuracy of 0.9320.}
\end{table}

\begin{figure}[t]
    \centering
    \includegraphics[scale=0.3, width=\linewidth]{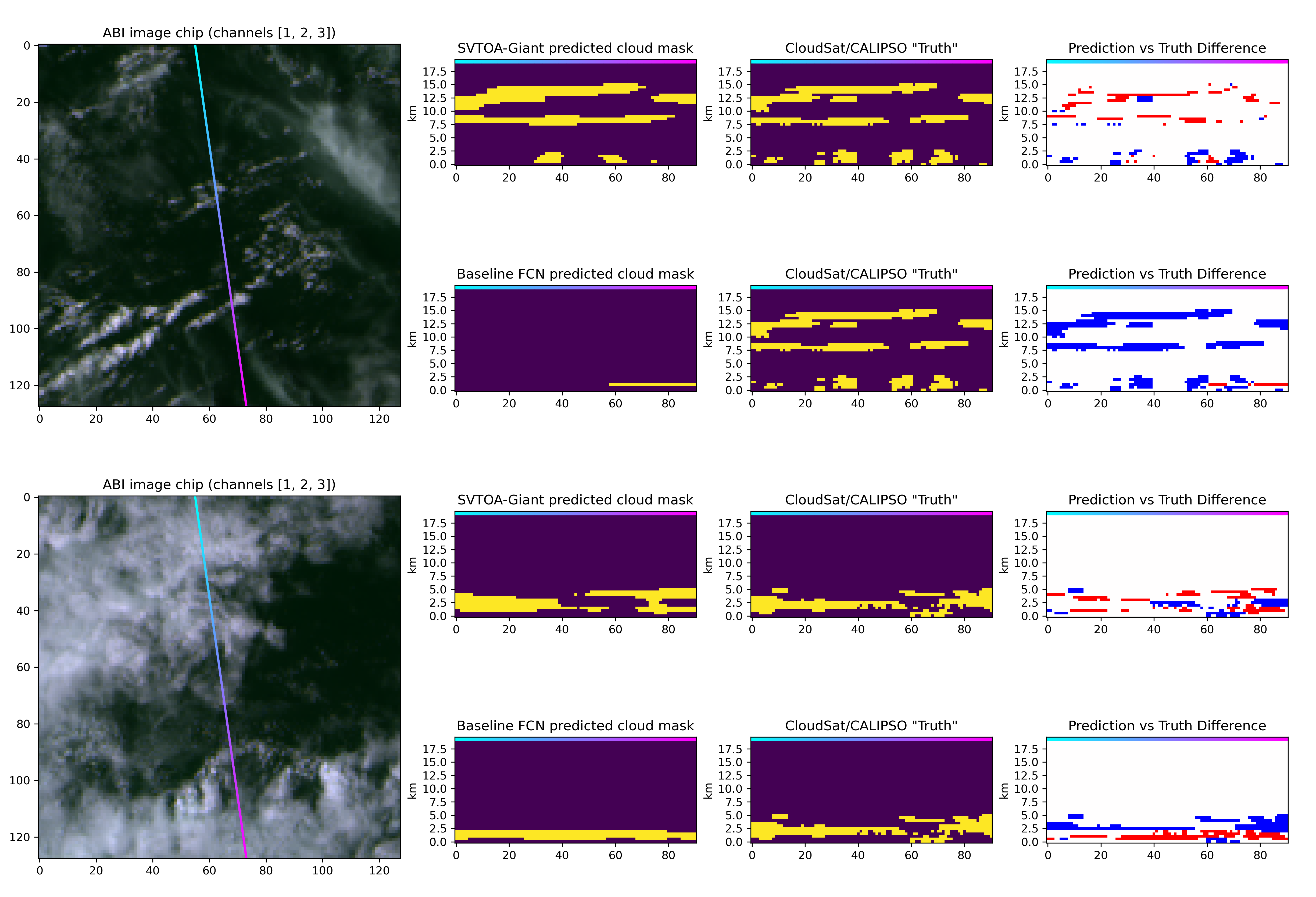}
    \caption{Examining model performance on 3D cloud retrieval downstream task. \textbf{Left}: ABI image chips showing the CloudSat/CALIPSO curtain transect. \textbf{Middle-left}: Cloud predictions from the SatVision-TOA-Giant and baseline FCN models. The bottom color bar indicates the location along the transect where clouds are predicted by the models. \textbf{Middle-right}: Ground truth vertical cloud retrieval mask from CloudSat/CALIPSO. \textbf{Right}: Difference between the predicted vertical cloud mask and the ground truth, where red indicates false positives and blue indicates false negatives.}
    \label{fig::viz}
\end{figure}

Cloud 3D structures dominate the impact to cloud radiative feedback at the top of the atmosphere \cite{shonk_effect_2010, li_radiative_2011} and influence precipitation initiation time, intensity and duration at the surface \cite{rajeevan2013study, gong_linkage_2020}. As such, the international GEWEX cloud assessment program recommends that cloud retrieval algorithms should focus on the vertical structure of clouds \cite{stubenrauch_assessment_2013}. Unfortunately, at present, operational cloud masks are still only reported at 2D levels for passive space-borne instruments, such as MODIS and GOES-ABI. These limited masks are then used in the decision tree for many other products that are obscured by clouds (e.g., aerosol, land surface Bidirectional Reflectance Distribution Function (BRDF), active fire, etc.) to prevent next-step retrievals from occurring when a cloud is reported for the pixel. This workflow results in a significant under-utilization of information for retrieving features below clouds. 

Passive imagers with visible and infrared channels (e.g., MODIS, ABI) sense the cloud top height and cloud optical thickness through the CO$_2$ slicing mechanism and multi-spectral reflectance, respectively \cite{platnick_modis_2017}. These are the fundamental physical basis that inspire several recent ML-based vertical cloud mask retrieval works, all using various standard (FCN, UNet) deep learning models \cite{noh_framework_2022, wang_toward_2023, foley_3-d_2024}. While overall successes are reported, these works also found ML is not a panacea to all situations. In particular, thin low clouds, deep convective clouds with cirrus above or thin boundary layer clouds beneath, and overlapping clouds, are reported to be extremely hard to predict accurately. Thin low clouds are the most prevailing cloud types, and they are the major cooling factor to combat global warming. Overlapping clouds occupy about 17\% of the global cloud population \cite{marchant_evaluation_2020}, and warm surface 
more compared to single-layer cloud with the same optical thickness and cloud top height \cite{li_radiative_2011}. However, Marchant et al. \cite{marchant_evaluation_2020} compared the MODIS Collection 6 overlapping cloud flag and collocated CloudSat radar + CALIPSO lidar retrievals, and found they only agreed 12\% of the time. 

In this downstream task of SatVision-TOA, the aim is to learn and predict the CloudSat+CALIPSO vertical cloud masks. We face two challenges. For one, we are applying a model pre-trained on MODIS TOA data to the ABI observations. Although the 14 ABI channels that best match MODIS channel frequencies are selected and ABI chips near the peripheries of the full-disk image are excluded considering strong slantwise-view distortion, we did not apply any further calibration nor image re-gridding to match MODIS footprint size. Moreover, ABI scans a fixed disk with 15 minutes refresh rate, covering the entire diurnal cycle, while MODIS-TOA model was trained on 9AM fixed local time TERRA-MODIS images with global coverage.
This means that some of the ABI images embedding strong cloud or precipitation diurnal cycles were never seen by SatVision-TOA before. The transfer learning skill is one sub-task we intentionally designed to evaluate the foundation model application range across similar but different instruments onboard different platforms. Therefore, outcomes of this downstream task will inform the applicability of this foundation model to other satellite measurements using MODIS-like instruments. These include but are not limited to NOAA’s VIIRS, NASA’s PACE-Ocean Color Instrument (OCI), Japan’s HIMAWARI-Advanced Himawari Imager (AHI), European’s METEOSAT-MTG (Meteosat Third Generation), and future NASA-NOAA collaborative Geostationary Extended Observation (GEO-XO) mission.

With the two challenges in-mind, we devised a relatively simple downstream task: only making predictions along the CloudSat-CALIPSO curtain. The training dataset is made of  128$\times$128 pixel ABI image chips from 14 channel measurements with the CloudSat/CALIPSO overpasses the center of the cropped image. Because CloudSat/CALIPSO is on a sun-synchronized orbit with equator passing time of 1:30AM/1:30PM, the selected training and testing samples are confined to 1:30PM local time because we want to use all 14 channel measurements. A total of 7000 ABI chips were used for training and another randomly selected 1300 chips are kept for independent validation. The baseline DL model used a FCN model structure.

We present results in Table \ref{tab:3dcloud_retrieval_table}. SatVision-TOA Giant pre-trained with 100 M image chips achieves the highest mIOU score of 0.4638, an accuracy of 95.7\%, and an Area Under the ROC Curve (AUC) score of 0.98 over the baseline FCN model which achieved a mIOU score of 0.2185, an accuracy of 93.2\%, and an AUC score of 0.93. In binary classification tasks, such as 3D cloud retrieval, accuracy can be misleading, particularly when class distribution is imbalanced, as it may inflate performance by favoring the majority class. Metrics such as mIOU offer a more reliable assessment by considering both false positives and false negatives. Additional results comparing different SatVision-TOA model configurations are presented in Appendix Table \ref{tab:3dcloud_retrieval_table_config}, \ref{tab:l1_loss}, and in Appendix \ref{fig::3dcloud_roc_curve}.

\section{Discussion}

SatVision-TOA represents a groundbreaking advancement in the use of deep learning for remote sensing applications. To our knowledge, SatVision-TOA is the largest (in terms of model parameters and number of training images) vision transformer-based foundation model developed exclusively with remotely sensed data. Trained on an unprecedented 100 million MODIS image chips spanning more than a decade and leveraging over 100 TB of data, SatVision-TOA sets a new benchmark for large-scale, moderate- to coarse-resolution deep learning models. Its design uniquely captures global-scale spatial relationships and temporal patterns while maintaining high data efficiency and adaptability across diverse downstream tasks. We evaluated SatVision-TOA by examining its performance in both image reconstruction and a downstream task (3D cloud retrieval), establishing the baseline for countless data-driven science applications powered by foundation models.

\subsection{Reconstruction Insights: Learning Generalizable Representations}

The image reconstruction experiment not only highlights the model’s ability to reproduce input data but also underscores its capacity to learn complex spatial patterns and transitional regions critical for tasks like land and cloud modeling. Achieving an exceptionally high SSIM score of 0.9289 on the test dataset, SatVision-TOA demonstrates its ability to reconstruct fine-grained details and maintain structural coherence, even across masked regions. Figure \ref{fig:viz} demonstrates three example images at different levels of complexity and their reconstruction and reports the SSIM score for each image. Visual results (Figure \ref{fig:viz}) reveal how the model effectively preserves the continuity of features spanning multiple patches, offering strong evidence of its ability to generalize across diverse geophysical conditions. The results demonstrate that SatVision-TOA pre-trained on 100 million image chips achieves a remarkably high SSIM score of 0.9289. These results clearly indicate that the model can generalize and reconstruct complex masked regions with strong fidelity to the original image chip while preserving fine-grained details and structural coherence. More so, SatVision-TOA is capable of seamlessly reconstructing regions with complex cloud dynamics which exemplifies the strong capabilities of this model across different types of landscape and atmospheric conditions. This capacity is critical to its extensibility in fine-tuning for specialized applications.

\subsection{Downstream Performance: Excelling in 3D Cloud Retrieval}

The primary value of foundation models is their ability to be "fine-tuned" for specific tasks (aka downstream tasks) with a minimal amount of labeled training data and compute time).  In the downstream 3D cloud retrieval task, SatVision-TOA dramatically outperforms a baseline model, such as FCN trained from scratch, with >50\% higher true positive cloud detection rates in challenging scenarios like boundary-layer thin clouds and multi-layer formations (Figure \ref{fig::viz} and Table \ref{tab:3dcloud_retrieval_table}). In Figure \ref{fig::viz} we show 2 examples corresponding to challenging situations mentioned previously, which are boundary layer thin clouds (Row. 1) and multi-layer clouds (Row. 2). Additional examples may be found in Appendix \ref{fig::3dcloud_app_0}.  To ensure fairness in our experiments, both the baseline model and the SatVision-TOA downstream task use the same FCN decoder architecture and labeled training data sets (n = 7000) in the training of the models.  As such, the difference in results is strictly due to the difference in the encoder backbone: the pre-trained SatVision-TOA backbone versus the from-scratch FCN encoder. 

Notably, SatVision-TOA more accurately captures broken cloud structures and cloud inhomogeneity, critical for realistic cloud modeling and digital twin applications. For optically thick clouds, it demonstrates superior performance in estimating cloud geometric thickness, and accurately predicting cloud top and bottom heights, even under complex atmospheric conditions (see Appendix \ref{fig::3dcloud_app_0}). Only daytime MODIS Terra images are used for training this model, hence, it is expected to have some limitations in downstream tasks that involve diurnal variations (e.g., cloud life cycle). Nevertheless, this advancement addresses long-standing limitations in cloud representation, particularly in multilayer scenarios, and bridges critical gaps in remote sensing for atmospheric research with limited data and training time for fine-tuning requirements.

\subsection{Versatility and Transferability: Bridging Modalities}

SatVision-TOA’s adaptability extends beyond MODIS data. By fine-tuning on GOES-ABI data with distinct spatial resolutions and spectral profiles, the model maintains high performance with minimal retraining. This transferability was evidenced by fine-tuning SatVision-TOA for a specific science question such as 3D cloud retrieval, achieving remarkable performance compared to its baseline FCN counterpart, without requiring extensive task-specific retraining. This adaptability, enabled by its vision transformer architecture, underscores its efficiency in transferring learned representations to new domains and data modalities. The efficiency of SatVision-TOA in fine-tuning on small datasets demonstrates the scalability and versatility of foundation models for computer vision science applications, paving the way for rapid deployment in diverse Earth observation tasks, including operational atmospheric monitoring and environmental research, largely reducing the costs of training data development and compute time.

\subsection{Redefining "All-Sky" Modeling}

A key innovation of SatVision-TOA lies in its inclusion of "all-sky" conditions during pre-training, incorporating a vast range of cloud conditions often excluded in traditional models. This intentional design allows the model to capture the structural and spectral variability of atmospheric layers, enabling it to tackle cloud-rich tasks critical to atmospheric sciences. Unlike models optimized solely for cloud-cleared data, SatVision-TOA addresses the challenges of natural atmospheric variability, unlocking new potential for applications such as cloud lifecycle studies and global climate monitoring. This approach redefines how deep learning models approach atmospheric phenomena, positioning SatVision-TOA as a foundational tool for addressing previously overlooked challenges in atmospheric research. This capability enables scientists at different skill levels and from different backgrounds to easily and in a very short order develop deep learning data-driven applications to answer some of the most complex questions using remote sensing data.

Overall, SatVision-TOA exemplifies a paradigm shift in the use of foundation models for Earth observation within medium- and coarse-resolution remote sensing data. Its ability to adapt across datasets, reconstruct fine-grained spatial features, and excel in downstream tasks showcases the transformative potential of large-scale pre-training on moderate- to coarse-resolution remote sensing data. By investing a significant amount of image pre-processing and compute to pre-train this model, SatVision-TOA will enable a countless set of opportunities for research that can leverage the feature representation and learning capabilities of SatVision-TOA to expand for their own science needs. Additionally, by addressing critical gaps in cloud characterization, cross-domain transferability, and data efficiency, SatVision-TOA sets a new standard for scalable, versatile models in atmospheric science and environmental monitoring. This research not only expands the boundaries of what is possible with deep learning in remote sensing but also paves the way for the next generation of data-driven insights into Earth system processes.

\section{Conclusion}
The full potential of deep learning-driven science applications leveraging moderate- to coarse-resolution data remains untapped. To address this, SatVision-TOA, a large-scale, open-source vision transformer model with 3 billion parameters, was trained on an unprecedented 100-million-image dataset derived from NASA MODIS TOA data, encompassing over 100 TB of data across a decade. By utilizing MODIS’s high temporal frequency and preserving spatial relationships through its transformer-based architecture, SatVision-TOA offers significant advancements in the remote sensing deep learning domain. Its innovations are encapsulated in three key points:

\begin{itemize}
  \item Enhancing Large-Scale Earth Observation Workflows: SatVision-TOA establishes an efficient workflow for generating massive training datasets and models, balancing spatial and temporal trade-offs to enable rapid development of machine learning applications with minimal training data.
  \item Adaptable and Scalable Model Design: SatVision-TOA outperforms traditional models like FCN in reconstructing complex cloud structures, unifying diverse Earth observation datasets, and reducing dependency on task-specific models.
  \item Foundation for Data-Driven Remote Sensing Applications: Leveraging one of the largest public datasets, SatVision-TOA captures global contexts and robust features, enabling transformative advancements in atmospheric science, cloud structure analysis, and Earth system modeling.
\end{itemize}

SatVision-TOA exemplifies a scalable, adaptable model designed to harness large archives of medium- and coarse-resolution data, addressing and closing critical gaps in training and inference for deep learning in Earth observation science applications. In the spirit of open-science we have released the weights and workflows in the hopes of broadening participation and fostering collaboration. These weights and workflows are available through HuggingFace and GitHub, respectively.\footnote{SatVision-TOA weights: https://huggingface.co/nasa-cisto-data-science-group/satvision-toa-giant-patch8-window8-128} \footnote{Associated code: https://github.com/nasa-nccs-hpda/pytorch-caney, DOI: https://doi.org/10.5281/zenodo.14206374}.

\section{Acknowledgments}
Computing resources supporting this work were provided by the NASA High-End Computing (HEC) Program through the NASA Center for Climate Simulation (NCCS) at Goddard Space Flight Center. This research used resources of the Oak Ridge Leadership Computing Facility at the Oak Ridge National Laboratory, which is supported by the Office of Science of the U.S. Department of Energy under Contract No. DE-AC05-00OR22725.. Downstream task example testing is partially supported by a NASA GSFC internal seed funding.

\newpage
\appendix
\section{Hardware and Software}
Our pre-training and data generation utilized NASA GSFC’s NCCS Discover Supercomputer and the Frontier Supercomputer at the Oak Ridge Leadership Facility. The SwinV2 Huge and SwinV2 Giant models pre-trained with 2 million and 26 million image chips utilized the Discover supercomputer. The SwinV2 Giant pre-trained with 100 million image chips utilized the Frontier supercomputer. The SCU17 partition of the Discover supercomputer has a 40-core AMD Rome CPU and four NVIDIA A100-40GB GPU accelerators. Each Frontier node has a 64-core AMD EPYC CPU with four AMD Instinct MI250X GPU accelerators. We used Pytorch \cite{paszke_pytorch_2019} and DeepSpeed \cite{rasley2020deepspeed} as the stack for performing pre-training.

Each fine-tuning experiment was conducted on 4 NVIDIA A100 GPUs with an input size of 128$\times$128$\times$14. An effective batch size of 8 was utilized. We employed the AdamW optimizer, couples with a linear warm up learning rate scheduler with a starting rate of $5e-7$, eventually peaking at $3e-4$.

Total compute spent pre-training the various model experiments was 68,329 GPU hours, or 7.8 GPU years.

\section{Pre-Training}
We use the fused AdamW \cite{loshchilov2017decoupled} optimizer with $\beta_1 =$ 0.9, and $\beta_2 =$ 0.99, with the OneCycle \cite{smith2019super} cosine learning rate scheduler, with the maximum learning rate variable between experiments due to scaling pre-training between the different dataset sizes. We utilized the SwinV2 Huge (658 Million parameters) and the SwinV2 Giant (3 Billion parameters) configurations of the SwinV2 model. For each pre-training experiment, we trained the models for 50 epochs. The input image size of the MODIS TOA image chips is 128$\times$128 with a total of 14 bands. The patch size for MIM pre-training was 8$\times$8 with a window size of 12. 

We employed bfloat16 precision \cite{kalamkar_study_2019} during pre-training, which optimized memory usage for both the model and the data. Bfloat16, a compact 16-bit floating-point format, retains the dynamic range of the standard 32-bit float (float32) but with reduced precision. This approach enabled more efficient use of hardware resources, allowing for the training of larger models or processing of larger datasets without compromising the model's ability to represent a wide range of values. By utilizing bfloat16, we effectively balanced computational efficiency with the need to maintain a sufficiently broad numerical range, ensuring robust model performance during pre-training. The bfloat16 format retains all the precision of the original remote sensing data.

\section{Model and Dataset Scaling Details}
In pre-training with the SwinV2 Giant and SwinV2 Huge models, we leveraged DeepSpeed’s Zero Redundancy Optimizer (ZeRO) distributed strategy techniques which were instrumental in managing the increased computational demands of the 658 million and 3 billion parameter vision models. 

In addition to scaling model size, we conducted experiments to scale dataset (batch) size. We observed that increasing the batch size while maintaining model performance was key to reducing the time for pre-training. We identified optimal batch sizes that balanced GPU memory usage without decreasing model performance. Scaling the learning rate to account for larger batch sizes allowed for continued model performance stability. We utilized the following scaling method. For the SwinV2 Giant pre-trained with 100 million chips, we found that scaling the learning rate to have a maximum of 0.00159727 allowed us to reach a maximum effective batch size of 16,384, spread across 256 GPUs on the Frontier supercomputer.

\section{Model Configuration Comparison} A key focus of our study is the relationship between both model and dataset scale and model performance on downstream tasks. We designed different experiments to test the robustness of model performance against input sample size and number of parameters.  To explore this, we pre-trained SwinV2 Giant models on three datasets of varying sizes- 2 million, 26 million and 100 million MODIS TOA image chips (subsets of larger images). This systematic comparison allows us to evaluate how increasing the volume of pre-training data impacts the model’s ability to generalize to specific tasks. By examining the gains and diminishing returns associated with larger datasets, we provide insights into the optimal scale of data needed for effective model training. Additionally, this approach aids in identifying the balance between computational resources used and model performance, which guides future efforts in visual foundation model development and deployment for remote sensing applications.

\newpage

\begin{table*}[ht]
    \centering
    \resizebox{\textwidth}{!}{
    \begin{tabular}{cc|cccccccccccccc}
    \Xhline{3\arrayrulewidth}
        Backbone & Pre-train Dataset & \multicolumn{14}{c}{SSIM Score per Channel} \\
        \Xhline{3\arrayrulewidth}
        &  & Ch1 & Ch2 & Ch3 & Ch6 & Ch7 & Ch21 & Ch26 & Ch27 & Ch28 & Ch29 & Ch30 & Ch31 & Ch32 & Ch33 \\\hline
        SVTOA-Huge & 2M & 0.89 & 0.87 & 0.90 & 0.85 & 0.86 & 0.90 & 0.92 & 0.94 & 0.94 & 0.92 & 0.89 & 0.92 & 0.92 & 0.93 \\\hline
        SVTOA-Giant & 2M & 0.89 & 0.88 & 0.91 & 0.86 & 0.86 & 0.91 & 0.91 & 0.95 & 0.95 & 0.92 & 0.92 & 0.92 & 0.92 & 0.93 \\\hline
        SVTOA-Giant & 26M & 0.91 & 0.89 & 0.92 & 0.88 & 0.88 & 0.92 & 0.95 & 0.95 & 0.87 & 0.93 & 0.90 & 0.93 & 0.93 & 0.91 \\\hline
        SVTOA-Giant & 100M & 0.91 & 0.90 & 0.93 & 0.89 & 0.89 & 0.93 & 0.95 & 0.95 & 0.95 & 0.94 & 0.94 & 0.94 & 0.94 & 0.95 \\\hline
    \Xhline{3\arrayrulewidth}
    \end{tabular}
    }
    \vspace{0.02in}
    \caption{\label{fig:ssim_channel} SSIM score per channel for different SwinV2 models and dataset sizes. The table presents SSIM scores across 14 channels for SwinV2-Huge and SwinV2-Giant model configurations with varying pre-training dataset sizes (2M, 26M, 100M image chips).}
\end{table*}

\begin{table}[t]
    \centering
    \resizebox{0.49\textwidth}{!}{
    \begin{tabular}{cccc}
    \Xhline{3\arrayrulewidth}
        Backbone & Model Size (parameters) & Pre-Train Dataset & SSIM Score \\
        \Xhline{3\arrayrulewidth}
        SVTOA-Huge & 658M & 2M & 0.9032 \\\hline
        SVTOA-Giant & 3B & 2M & 0.9084 \\\hline
        SVTOA-Giant & 3B & 26M & 0.9115 \\\hline
        SVTOA-Giant & 3B & 100M & 0.9289 \\\hline
        \Xhline{3\arrayrulewidth}
    \end{tabular}
    }
    \caption{\label{tab:ssim_score_configs} Structural Similarity Index Measure (SSIM) score comparisons for various SwinV2 model sizes and pre-training dataset sizes. Higher SSIM scores indicate better structural similarity in model predictions relative to ground truth, showcasing the impact of both model scale and the volume of pre-training data on performance.}
\end{table}

\begin{table}[t]
    \centering
    \resizebox{0.49\textwidth}{!}{
    \begin{tabular}{cccc}
    \Xhline{3\arrayrulewidth}
        Backbone & Pre-Train Dataset & mIOU & Accuracy \\
        \Xhline{3\arrayrulewidth}
        FCN Baseline & & 0.2185 & 0.9320\\\hline
        SVTOA-Huge & 2M & 0.3838 & 0.9510\\\hline
        SVTOA-Giant & 2M & 0.3695 & 0.9492\\\hline
        SVTOA-Giant & 26M & 0.4144 & 0.9543\\\hline
        SVTOA-Giant & 100M & 0.4638 & 0.9574\\\hline
        \Xhline{3\arrayrulewidth}
    \end{tabular}
    }
    \caption{\label{tab:3dcloud_retrieval_table_config} 3D cloud retrieval results for multiple model and pre-training dataset size configurations.}
\end{table}

\begin{table}[t]
    \centering
    \resizebox{0.49\textwidth}{!}{
    \begin{tabular}{cccc}
    \Xhline{3\arrayrulewidth}
        Backbone & Model Size (parameters) & Pre-Train Dataset & l1 Loss \\
        \Xhline{3\arrayrulewidth}
        SVTOA-Huge & 658M & 2M & 0.0208 \\\hline
        SVTOA-Giant & 3B & 2M & 0.0204 \\\hline
        SVTOA-Giant & 3B & 26M & 0.0207 \\\hline
        SVTOA-Giant & 3B & 100M & 0.0173 \\\hline
        \Xhline{3\arrayrulewidth}
    \end{tabular}
    }
    \caption{\label{tab:l1_loss} L1 loss values for various combinations of SatVision-TOA model configurations (Giant, Huge) and dataset sizes (2M, 26M, and 100M image chips). Lower L1 loss values indicate better model accuracy in predicting pixel-level intensities.}
\end{table}

\newpage

\begin{figure}[t]
    \centering
    \includegraphics[scale=0.2, width=\linewidth]{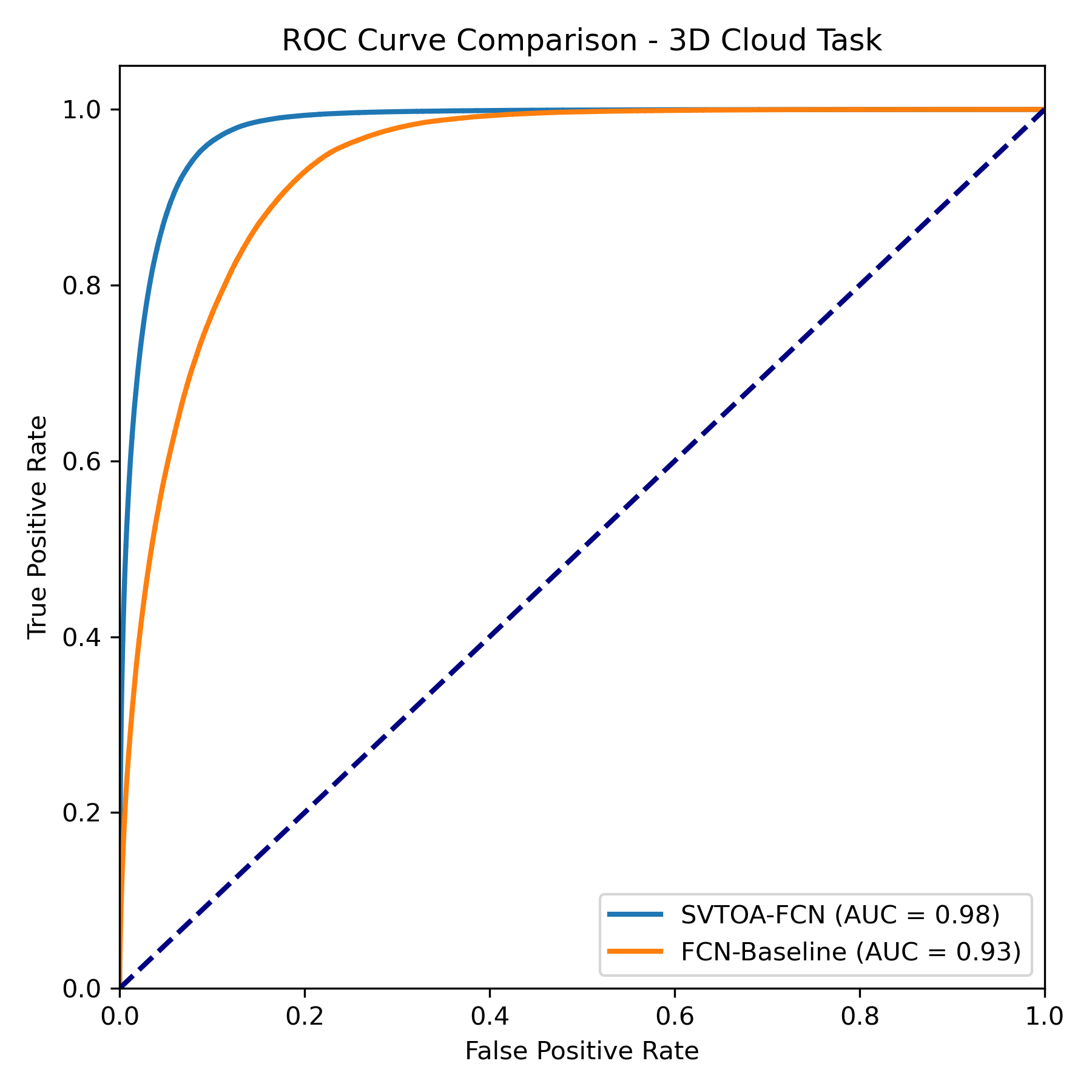}
    \label{fig::3dcloud_roc_curve}
    \caption{3D Cloud Retrieval Results: Receiver Operating Characteristic (ROC) Curve Comparison, SatVision-TOA (SVTOA-FCN) vs Baseline (FCN-Baseline) models- Highlighting performance differences in classification accuracy with Area Under the Curve (AUC) values for each model.}
\end{figure}

\begin{figure}[t]
    \centering
    \includegraphics[scale=0.1, width=\linewidth]{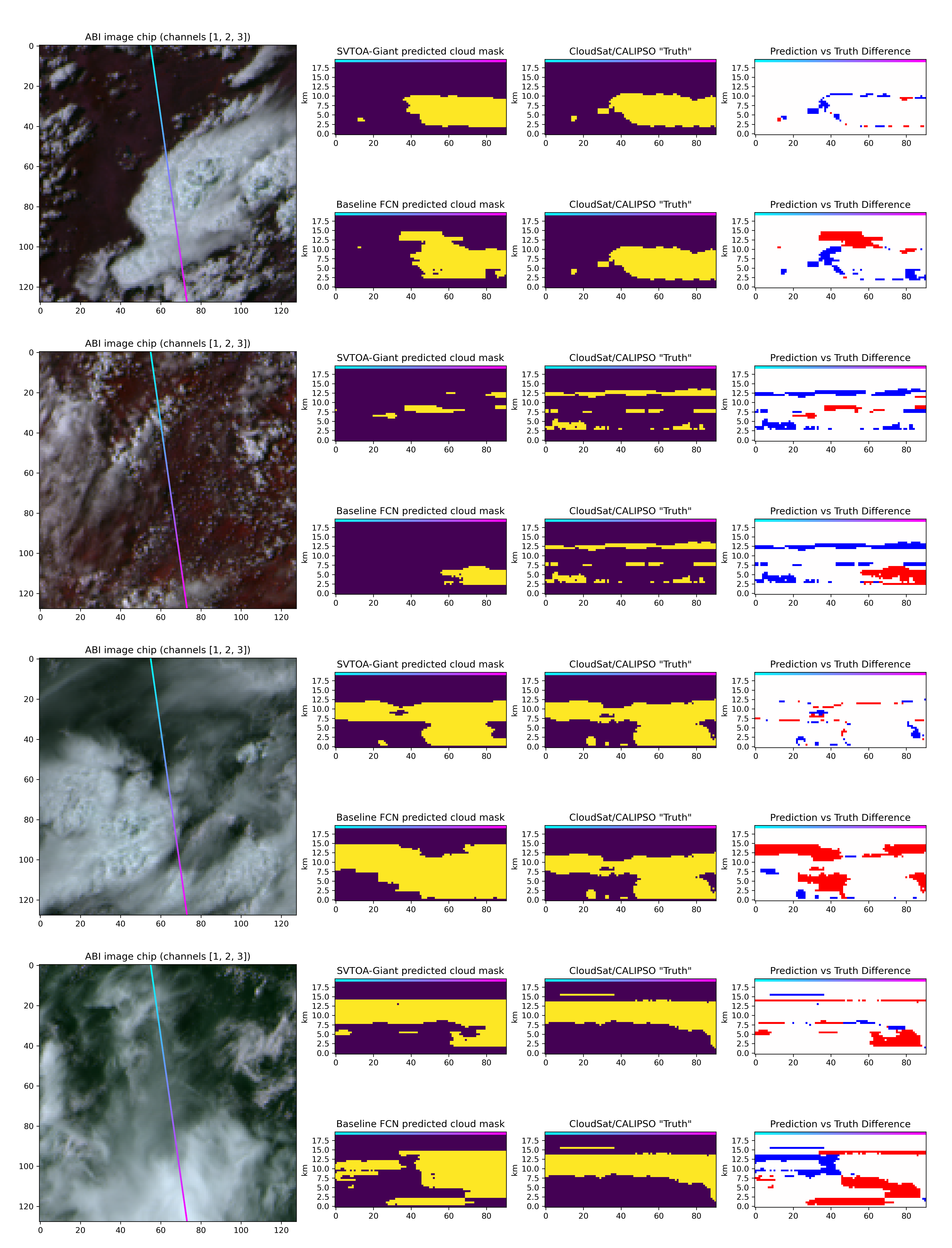}
    \label{fig::3dcloud_app_0}
    \caption{3D Cloud Retrieval Predictions}
\end{figure}

\begin{figure}[t]
    \centering
    \includegraphics[scale=0.1, width=\linewidth]{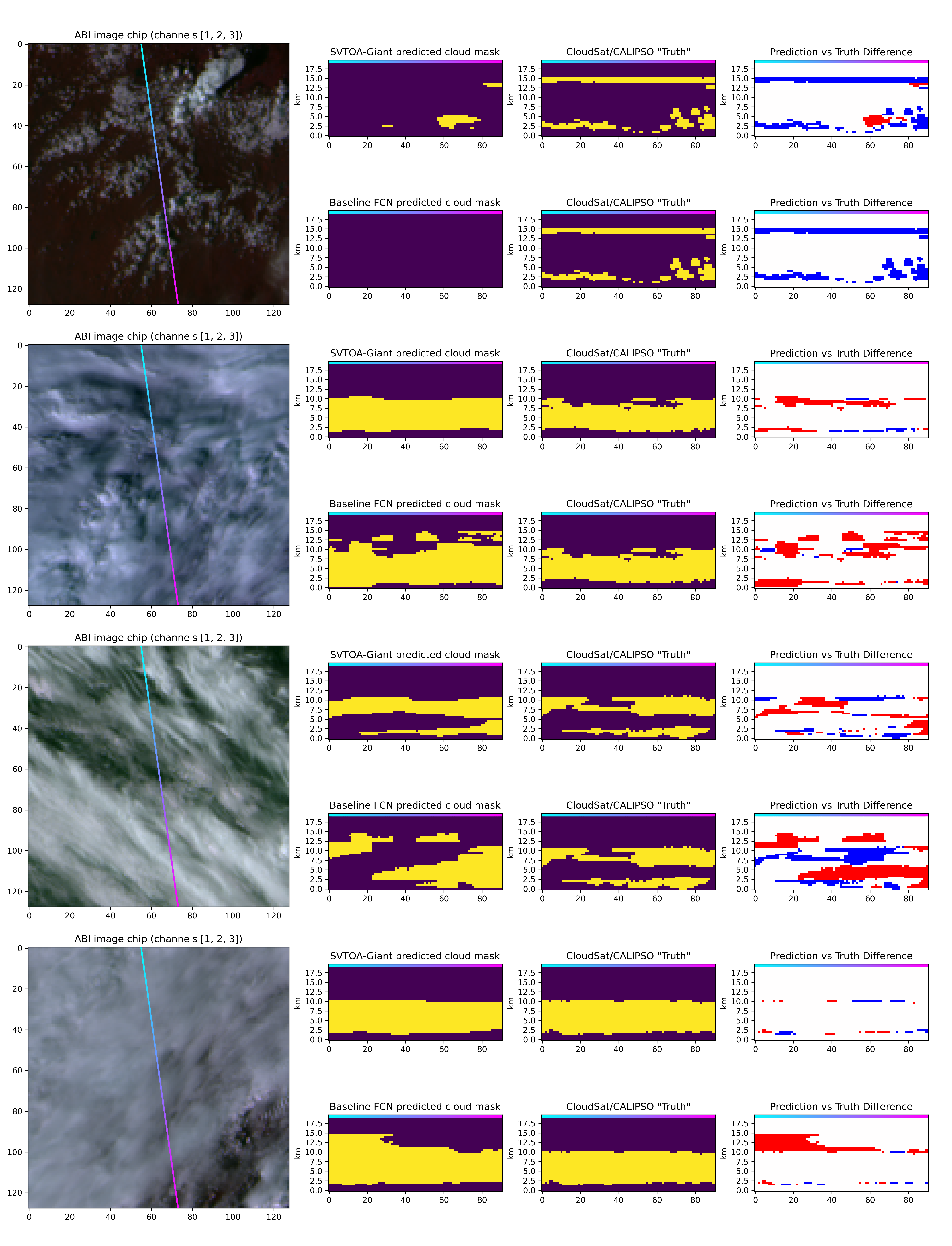}
    \label{fig::3dcloud_app_1}
    \caption{3D Cloud Retrieval Predictions}
\end{figure}

\end{document}